# Possibilistic decreasing persistence

**Dimiter Driankov**
Department of Computer and Information Science
Universitetet i Linköping
S-58183 Linköping
Sweden
email: ddr@ida.liu.se

**Jérôme Lang** *
IRIT
Université Paul Sabatier
F-31062 Toulouse Cedex
France
email: lang@irit.fr

## Abstract

A key issue in the handling of temporal data is the treatment of persistence; in most approaches it consists in inferring defeasible conlusions by extrapolating from the actual knowledge of the history of the world; we propose here a *gradual* modelling of persistence, following the idea that persistence is *decreasing* (the further we are from the last time point where a fluent is known to be true, the less certainly true the fluent is); it is based on possibility theory, which has strong relations with other well-known ordering-based approaches to nonmonotonic reasoning. We compare our approach with Dean and Kanazawa's probabilistic projection. We give a formal modelling of the decreasing persistence problem. Lastly, we show how to infer nonmonotonic conclusions using the principle of decreasing persistence.

## 1 Introduction

The use of persistence in order to draw nonmonotonic conclusions has been widely studied. Most approaches select models having the minimal set of changing fluents. Thus, in these approaches, a propositional fluent $f$ true at a given time point will tend to remain true indefinitely, provided that no other proposition being contradictory with $f$ is observed at a later time point; this is an extremely *adventurous* choice, and it may be often unrealistic, because some fluents have only a limited tendency to persist (for instance, given that it is raining at $t_0$, it is not reasonable to infer that it is still certainly raining one week later). Let us now consider a second typical case, where a fluent $f$ is known to be true at time $t_0$ and known to be false at a later time point $t_1$, nothing being known inbetween (for instance, it is raining at 10 am, and it is not raining at 6 pm). In this figure, there must be a time point $t*$ in $(t_0, t_1)$ when $f$ changes its truth value from true to false (this is known as the *clipping problem*). Chronological minimization (Shoham 88) and similar approaches prefer models where fluents change at the latest possible time point; this has been argued as being often unreasonable (see (Sandewall 92) for a discussion) and several other approaches have been proposed which reject the latter principle, and, cautiously, do not conclude anything about $f$ within $(t_0, t_1)$. For instance, the logic for time of action proposed in (Sandewall 92) will conclude that the truth value of $f$ is occluded during $(t_0, t_1)$. Borillo & Gaume's (90) three-valued extension of Kowalski & Sergot's event calculus will also give a cautious result. We argue that these cautious results assuming complete ignorance within the whole interval are not always realistic, since we are not always completely ignorant of what happens at time points being very close to one of the bounds of the interval (thus, in our example it is rather sure it is still raining at 10.05 am and rather sure it is not raining at 5.55 pm). The transition model given in (Cordier & Siegel 92) enables to specify explicitely whether fluents tend to persist or not depending on some applications conditions, and has thus a rich expression power, but however it cannot express that persistence may decrease gradually.

The reason why all these approaches cannot model decreasing persistence is clearly their *lack of graduality*; consider again the first *raining* example (forward projection); one is likely to believe that *rains* is almost certainly true a short time after 10.00, and not to believe anything at all after a very long time (say, one week later); note that in this latter case $\neg f$ should not be believed either; we are too far from a time-point when the truth value of $f$ is known for assuming anything: we are thus in a state of *complete ignorance* about the truth value of $f$. Between these two extreme states of knowledge, there is a lot of intermediary states, since the further from 10.00, the less certain we are that it is still raining: as time goes on, the amount of ignorance increases. This principle will be called **increasing ignorance about persistence**, or, for the sake of brevity, **decreasing persistence**, although we prefer the former formulation: indeed, what is gradually decreasing is not persistence of truth but persistence of our belief about truth; see (Asher 93) for a study of persistence of truth vs. persistence of belief. This *graduality* in persistence can be expressed in a qualitative way using ordering relations or in a more quantitative way using numerical measures of uncertainty.

---

*Much of this work was done while this author was visiting Linköping University

To our knowledge, there has been essentially one approach to modelling persistence in a gradual way, namely Dean and Kanazawa's *probabilistic projection* (Dean & Kanazawa 89a,b) (see also (Haddawy 90) for a temporal probability logic for reasoning about actions). They distinguish between 2 kinds of propositions, namely *facts* (or *fluents*) and *events*; a *fact* is a proposition which, once true, tends to persist, i.e. to remain true for some time without additional effort; *events* are instantaneous, and they do not persist, but they tend to change the truth value of some fluents. Note that all facts have a starting point and an ending point (possibly infinite); if a fluent is true, becomes false and then becomes true again, it must be considered as two different instances ("tokens") of the same fact. Dean and Kanazawa propose an elaborate probabilistic model for persistence, taking account for each fact of its natural tendency to persist, represented by a *survivor* function $S(\delta) = p(holds(f,t)|holds(f,t-\delta))$ (probability that $f$ survives at least for $\delta$ time units), and of the probabilities of events changing the truth value of the fluent. Thus probabilistic prediction comes down to computing the probability of $f$ being still true at $t$, or equivalently, the density function of the *clipping point* of $f$, i.e. the time point when it becomes false.

However, probabilistic prediction is not well-suited to dealing with fluents which may change their value several times; besides, a probabilistic modelling of persistence does not express that we become more and more ignorant about the truth value of a formula when time goes on. Let us consider the following example, where we know that it is raining at time $t_0$ (and that we do not know anything about what will happen afterwards). Dean and Kanazawa's approach will conclude that the probability of "raining" at $t_0 + \epsilon$ is close to 1 if $\epsilon$ is close to 0, which is intended; however, it will also conclude that if we are very far from $t_0$, *raining* is false, which is of course not intended. A first idea for treating this case correctly would be to model persistence with an asymptotic probability (which is actually the probability a priori that it is raining, independently from earlier and later observations); but it still does not express *increasing ignorance*, since probability theory is well-suited to modelling *chance*, but can not deal correctly with *ignorance* (see (Dubois & Prade 88)); possibility theory (Zadeh 78) is much more adapted to the representation of states of partial or complete ignorance.

A last point is that Dean and Kanazawa's probabilistic projection is only done forwards; our possibilistic approach also deals with *backwards* projection problems,
and also with *bounded projection* (see Section 3). After recalling the bases of possibility theory and its use in nonmonotonic reasoning, we will give a formal presentation of our approach, and lastly we will show how to use decreasing persistence in order to infer nonmonotonic conclusions.

## 2   Background on possibilistic logic

Let L be a classical propositional language (where ⊤ and ⊥ denote tautology and contradiction, respectively) and $\Omega$ be the classical set of interpretations associated with L. A *possibility distribution* is a mapping $\pi$ from $\Omega$ to $[0,1]$. $\pi$ is said to be *normalized* iff $\exists \omega \in \Omega$ such that $\pi(\omega) = 1$. By convention, $\pi$ represents some background knowledge about where the real world is; in particular, $\pi(\omega) = 0$ means that $\omega$ is not possible, and $\pi(\omega) = 1$ that nothing prevents $\omega$ from being the real world. When $\pi(\omega) > \pi(\omega')$, $\omega$ is a preferred candidate to $\omega'$ for being the real world. A possibility distribution leads to evaluate induces two mappings on L, namely a possibility measure $\Pi(\varphi) = Sup_{\omega \models \varphi} \pi(\omega)$ which evaluates the extent to which $\varphi$ is consistent with the available knowledge expressed by $\pi$, and a necessity (or certainty) measure $N(\varphi) = Inf_{\omega \models \neg\varphi}(1 - \pi(\omega)) = 1 - \Pi(\neg\varphi)$, which evaluates the extent to which $\varphi$ is entailed by the available knowledge. We have $\forall \varphi, \psi$, $N(\varphi \wedge \psi) = min(N(\varphi), N(\psi))$. Note that while $N(\varphi) = 1$ means that $\varphi$ is certainly true, $N(\varphi) = 0$ means only that $\varphi$ is not certain at all. Complete ignorance about $\varphi$ is expressed by $N(\varphi) = N(\neg\varphi) = 0$. Since possibility distributions are not required to be normalized, it may be the case that $N(\bot) > 0$. Note that we have $\forall \varphi, min(N(\varphi), N(\neg\varphi)) = N(\bot)$. Note that what is essential in possibility theory is not the precise value of certainty degrees, but their *ordinal* nature: indeed certainty degrees can be used to rank formulas of L. Namely, it is equivalent to work with necessity measures or with (qualitative) necessity relations (see (Dubois, Prade 91)) defined by $\geq_N$ defined by $\forall \varphi, \psi, \varphi \geq_N \psi$ iff $N(\varphi) \geq N(\psi)$, meaning that $\varphi$ is at least as certain as $\psi$.

A *possibilistic knowledge base* (Dubois et al. 91a) is a finite set of *necessity-valued formulas* $K = \{(\varphi_i \; \alpha_i), i = 1, n\}$ where $\alpha_i$ represents a lower bound of the necessity degree $N(\varphi_i)$. A possibility distribution $\pi$ on $\Omega$ satisfies $K$ iff $\forall i, N(\varphi_i) \geq \alpha_i$, where $N$ is the necessity measure induced by $\pi$. Logical consequence is then defined by $K \models (\xi \; \beta)$ iff any possibility distribution satisfying $K$ satisfies $(\xi \; \beta)$. The fuzzy set of models of a possibilistic knowledge base has for membership function the least specific possibility distribution satisfying the constraints $N(\varphi_i) \geq \alpha_i$, $i = 1, n$. This possibility distribution $\pi_K^*$ is defined by: $\forall \omega \in \Omega$, $\pi_K^*(\omega) = min_{i=1,n}\{1 - \alpha_i, \omega \models \neg\varphi_i\}$. Possibilistic logic allows for *partial inconsistency*, occuring there is no *normalized* possibility distribution satisfying $K$, which means that $K \models (\bot \; \beta)$ for some strictly positive $\beta$. The quantity $Max\{\beta, K \models (\bot \beta)\}$ is called *inconsistency degree* of $K$, denoted by $Incons(K)$. It can be shown that $Incons(K) = N_K^*(\bot) = 1 - Sup_{\omega \in \Omega} \pi_K^*(\omega)$.

In (Dubois et al. 91b), possibilistic logic was extended to a *timed* version which handles both uncertainty and time; basically, a *timed possibilistic knowledge base* consists in a collection of possibilistic knowledge bases indexed by time points varying on a given time scale T; so, instead of considering possibility distributions (resp. necessity measures), we consider collections of possibility distributions $\{\pi_t, t \in \mathcal{T}\}$ (resp. collections of necessity measures $\{N_t, t \in \mathcal{T}\}$).

From a possibilistic knowledge base $K$, it is possible to define a nonmonotonic inference relation (see (Dubois Prade 91)) $\vdash_K$ by: $\varphi \vdash_K \psi$ iff $N_K^*(\varphi \to \psi) > N_K^*(\neg\varphi)$. Note that in the particular case where $\varphi = \top$, we get the following (abbreviating $\top \vdash_t \psi$ in $\vdash_t \psi$): $\vdash \psi$ iff $N_K^*(\psi) > N_K^*(\bot)$ iff $N_t^*(\psi) > Incons(K)$. It has been shown that $\vdash_K$ enjoys all "desirable" properties that nonmonotonic inference relations "should" satisfy, including rational monotonicity (Dubois, Prade 91).

## 3 Possibilistic decreasing persistence: the extrapolation problem

### 3.1 Informal presentation of the extrapolation problem

The general principle of decreasing persistence is, given a factual temporal knowledge base and some information about the persistence of some given fluent $f$, to derive uncertain information about $f$ in the intervals when the truth value of $f$ is unknown. Let us start with motivating examples.

*Example 1* (unbounded forward extrapolation): let us consider the fluent $free$ of a given parking place which may or may not be free at any time-point $t$. Suppose that all we know about $free$ is that it holds up to $t_0 = 10.00$ (and we do not anything about it afterwards). We would like to extrapolate, using some knowledge describing how our ignorance about the persistence of $free$ increases, the following uncertain facts: the certainty (necessity) degree of $free$, which is 1 at 10.00 (since $free$ is known to be true), should be close to 1 when $t$ is close to 10.00 (we recall that $N(free)$ expresses to which point $free$ is entailed by the knowledge of reference; here it is obvious that at time points close to 10.00, $free$ is entailed, to some certainty degree close to 1, by both the fact that it holds at 10.00 and the general principle of decreasing persistence); then, the further $t$ is from 10.00, the less certain we are that $free$ is true; and there should also be a point from which on we are too far from 10.00 to be even weakly certain that $free$ still holds, i.e. from which on $N(free) = 0$ (then we are in a state of complete ignorance about $free$, i.e. we have $N(\neg free) = 0$ too). So, in this example the principle of decreasing persistence consists in extrapolating $N(free)$ in the interval $(10.00, +\infty)$; an example of persistence function is shown on Figure 1.

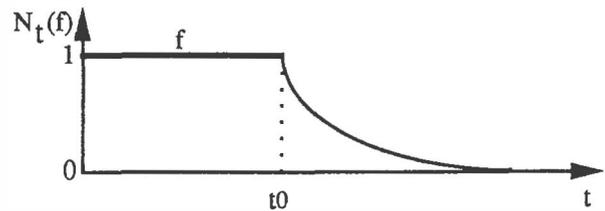

Figure 1: unbounded forward extrapolation

*Example 2* (unbounded backward extrapolation): assume now that $free$ is known to be true from 10.00 on (we do not know anything about it before) and we have to infer uncertain facts about the past of the fluent (this problem is also called *postdiction*). This case is very similar to forward extrapolation (in a symmetric way), and all previous remarks hold.

*Example 3* (bounded extrapolation without change): now, assume that $free$ is known to be true up to 10.00, and from 10.30 on, nothing is known about $free$ during the interval $(10.00, 10.30)$. Traditional nongradual approaches to persistence are too optimistic since they conclude by default that $free$ holds everywhere in $[10.00, 10.30]$, since nothing tells us that a change ocurred. However this is not always realistic, especially if the considered interval is long (relatively to the considered fluent). The most intuitive kind of extrapolation on $[10.00, 10.30]$ tells that the further from one of the two reference time-points 10.00 and 10.30, the less certain we are that $free$ still holds (see figure 2). The fact that $free$ holds at the two extremities of the interval should be a confirmation that $free$ holds in any arbitrary point of the interval; in other words, for instance, we should be at least as certain that $free$ holds at 10.15 in this situation than in the situation of Example 1. In some cases, the interval length may be too long for us to be somewhat certain that the fluent does not change within the interval; for instance, consider $free$ within $[10.00, 18.00]$. See Figure 2.

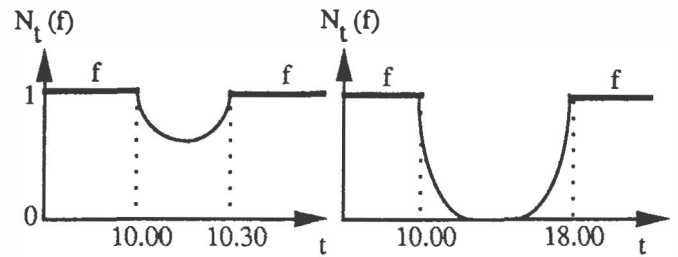

Figure 2: bounded extrapolation without change

*Example 4* (bounded extrapolation with change): now, assume $free$ is true up to 10.00, and false from 10.30 on; again, nothing is known during $(10.00, 10.30)$. Traditional non-gradual approaches are too cautious since they conclude that $free$ is unknown within (10.00, 10.30); however, a more realistic (and more informative) extrapolation would tell that *free* is rather certainly true if we are very close to 10.00 (the closer, the more certain; but it should nevertheless decrease faster than in Examples 1 and 3), and rather certainly false if we are very close to 10.30 (again, the closer, the more certain). See Figure 3.

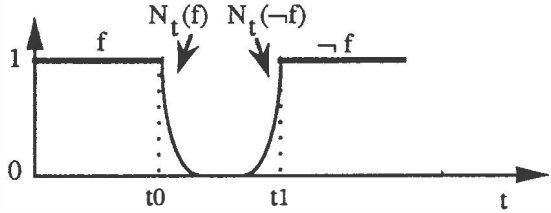

Figure 3: bounded extrapolation with change

## 3.2 Formalizing possibilistic decreasing persistence

First, it is primordial to state the distinction between *factual knowledge* and *knowledge about persistence*. The first one expresses what we know about the world during the time scale of reference and enables us only to draw certain, monotonic conclusions (for instance "it was raining from 10.00 to 11.00, and it was not raining at 12.30"), while the second one expresses what we know about the general behaviour of fluents (for instance, "*raining* tends to persist but usually no more than a couple of hours") and, together with factual knowledge, enables us to draw uncertain and defeasible conclusions.

### 3.2.1 Factual knowledge

*Factual knowledge* consists in an generally incomplete knowledge about the the world at every time point. It will be represented in a traditional way, by reifying time. Let $\mathcal{T} = (-\infty, +\infty)$ be the time scale of reference. Let L be a propositional logical language; atomic propositions which are allowed to vary along time are called *fluents*. A *timed knowledge base* $K$ is a finite set of *timed formulas* $\tau : \varphi$, where $\tau$ is a subset of T (generally an interval) and $\varphi$ a well-formed formula of L. $\tau : \varphi$ expresses that $\varphi$ holds for any time point $t$ in $\tau$. The *cut of $K$ at $t_0$* is the classical knowledge base $K_{t_0} = \{\tau : \varphi \in K \mid t_0 \in K\}$; clearly, a formula $\varphi$ is known to be true at $t_0$ iff $\varphi \in Cn(K_{t_0})$, where $Cn$ denotes logical closure, and known to be false at $t_0$ iff $\neg\varphi \in Cn(K_{t_0})$; if $\varphi$ is neither True nor False at $t_0$ then $\varphi$ is said to be *unknown* at $t_0$. Note that there is a fourth possible status for $\varphi$ at $t_0$, due to the possibility that $K_{t_0}$ be inconsistent (in which case $\varphi$ is both True and False); note that the set { True, False, Unknown, Inconsistent } is the well-known 4-valued lattice of (Belnap 77). However, for the sake of clarity, in this paper we will deliberately ignore inconsistent time-points (i.e. time-points $t$ such that $K_t$ is inconsistent), either by assuming that the timed knowledge base is maintained consistent, or by considering all contingent formulas as Unknown at inconsistent time-points.

The *partial history $H$* induced by $K$ is the logical closure of $K$, i.e. the collection of all $Cn(K_t)$, for $t$ varying in T. We will denote the belief status (True, False or Unknown) of $\varphi$ at $t_0$ by $H_t(\varphi)$.

### 3.2.2 Persistence extrapolation problems

Let $f$ be a propositional fluent, and let $H$ be a partial history on the time scale T. A time-point $t$ will said to be *informative* for $f$ iff $H_t(f) =$ True or $H_t(f) =$ False. The set of all informative time-points of $f$ is denoted by $ITP(f)$. For practical reasons we need to require that partial histories satisfy the following property: $H$ is said to be *closed* iff for any elementary fluent $f$, $ITP(\varphi)$ is a closed subset of T, i.e. a (possibly infinite) union of intervals of T which have one of these 4 forms: $[a,b]$ (possibly a = b), $[a,+\infty)$, $(-\infty, b]$ or $(-\infty, +\infty)$. $H$ being a closed partial history, a time-point $t$ is said to be a *reference time-point* for $f$ w.r.t. $H$ iff $t$ is at the leftest or at the rightest extremity of one of the intervals constituting $ITP(f)$. The complementary of $ITP(f)$, i.e. the set of all time points $t$ when $H_t(f) =$ Unknown, is a (possibly infinite) union of airwise disjoint open intervals, called *maximal non-informative intervals* of $f$ w.r.t. $H$; if $ITP(f) \neq \emptyset$, their form is either $(-\infty, t_0)$ or $(t_n, +\infty)$ or $(t_i, t_{i+1})$, where all $t_i$'s are reference time points for $f$ w.r.t. $H$ (it may be the case that $t_i = t_{i+1}$). From now on we exclude the trivial case $ITP(f) = \emptyset$ (i.e. the truth value of $f$ is always unkwown) since it is completely uninteresting (persistence cannot apply).

A *persistence extrapolation problem* consists in a closed history $H$, an elementary fluent $f$ and a maximal non-informative interval $I$ for $f$ w.r.t. $H$. The various examples presented informally in Section 3.1 suggest the following classification of persistence extrapolation problems:

- a persistence extrapolation problem $(H, f, I)$ is an *unbounded extrapolation problem* iff $I = (t_n, +\infty)$ (*forward extrapolation*), or $I = (-\infty, t_0)$ (*backward extrapolation*).

- a persistence extrapolation problem $(H, f, I)$ is a *bounded extrapolation problem without change* iff $I = (t_i, t_{i+1})$ and $H_{t_i}(f) = H_{t_{i+1}}(f)$.

- a persistence extrapolation problem $(H, f, I)$ is a *bounded extrapolation problem with change* iff $I = (t_i, t_{i+1})$ and $H_{t_i}(f) \neq H_{t_{i+1}}(\varphi)$.

### 3.2.3 Decreasing persistence functions and decreasing persistence schemata for fluents

Having stated persistence extrapolation problems, we are now giving a general methodology for solving them.

Informally, extrapolation based on decreasing persistence consists in inferring by default a truth-value, with some certainty degree, to a fluent at time-points where its truth-value is not definitely known. Of course, the way to cope with it may depend not only on the involved fluent, but on the class (backward, forward, ...) of the extrapolation problem and when it occurs. Let $I$ be a maximal non-informative interval for $f$ w.r.t. $H$. A *persistence function* for $(f, I)$ is a mapping from $I$ to $[0, 1]$ which associates to any $t$ in $I$ the necessity degree $N_t(f)$ of $f$ at t. Thus, persistence functions extrapolate uncertain knowledge from factual knowledge by using the general principle of decreasing persistence. Obviously, the problem is tractable only if the user can specify persistence functions in a general way (for instance, "in a forward extrapolation problem starting at $t_0$ the necessity degree of *free* decreases linearly and reaches 0 at $t_0 + 1.00$ if $t_0$ is during the day and at $t_0 + 4.00$ if $t_0$ is during the night"). This is a *decreasing persistence schema*. Once applied to a given partial history, a persistence schema is "instanciated" to persistence functions. If $H$ is a partial history and $Pers$ denotes a set of persistence schemata for a subset of the fluents involved in $H$, then $Apply(Pers, H)$ denotes the application of $Pers$ to $H$. Note that $Apply(Pers, H)$ is a collection of possibilistic knowledge bases (one for each $t$, denoted by $Apply(Pers, H)_t$). In next Section we investigate some of the properties that persistence schemata should preferably satisfy in order to be in accordance with the general principle of decreasing pereistence, and we propose some examples of persistence schemata.

## 4 From qualitative to quantitative axioms for persistence schemata

Independently from the exact shape of the persistence function of a fluent $f$ im an interval $I$, there are some very general properties that is may be desirable to impose. We give a first set of very basic axioms which are completely qualitative (since they do not use the metric nature of T and $[0, 1]$); we propose then a second set of more debatable properties, which are qualitative with respect to necessity degrees but quantitative with respect to time.

### 4.1 Basic axioms for persistence functions

These very basic axioms just ensure that persistence is well respecting the principle of *increasing ignorance*.

*Forward extrapolation*
Let $(H, f, (t_0, +\infty))$ be a forward extrapolation problem.
**D1**. $N_t(f)$ is non-increasing on $(t_0, +\infty)$
Obviously, D1 does not restrict a lot the possible persistence functions; typical examples of functions satisfying D1 are shown in Figure 4. But, however basic it is, D1 should sometimes not be required (for instance, for periodic or "usually periodic" fluents with a known period, like "sleep").

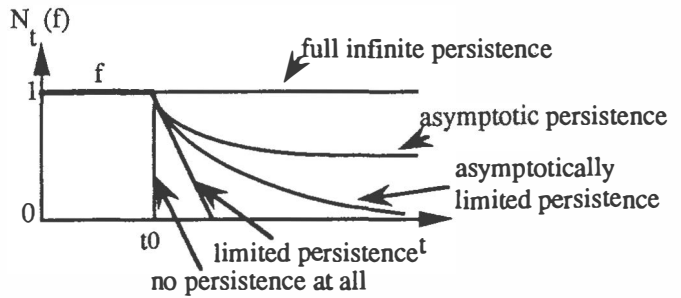

Figure 4: some forward persistence functions

On Figure 4 we have represented *continuous* functions satisfying D1 (except *no persistence at all*); note that any persistence function satisfying D1 and continuity is of one of the four following types shown on figure 4. Among other possible requirements, one could require the persistence function to be strictly decreasing on $[t_0, +\infty)$ (which rules out limited persistence functions) or, which is weaker, strictly decreasing in the right neighbourhood of $t_0$.
These requirements can be formulated in very similar ways for all other classes of extrapolation problems (for the sake of brevity we will omit doing it).

*Backward extrapolation*
This is very analogous to the case of forward persistence, except that persistence is "increasing" (but, of course, still decreasing with respect to the distance to the nearest reference time point): given a backward extrapolation problem $(H, f, (-\infty, t_0))$:
**D2**. $N_t(f)$ is non-decreasing on $(-\infty, t_0]$

*Bounded extrapolation without change*
Let $(H, f, (t_0, t_1))$ be a bounded extrapolation problem without change (without loss of generality, $f$ being True at both $t_0$ and $t_1$).
**D3**. $\exists t* \in [t_0, t_1]$ such that $N_t(f)$ is is non-increasing in $[t_0, t*]$ and non-decreasing in $[t*, t_1]$.
Strictness in the neighbourhoods of $t_0$ and $t_1$ would ensure that $t* \in (t_0, t_1)$). Note that the persistence function needs not to be symmetrical. Some admissible functions are shown on figure 5.

When the persistence function is continuous, it is necessarily of one of the 3 following types, shown on figure 5, depending on the minimal value of $N_t(f)$ on $[t_0, t_1]$: *full persistence*, where $\forall t \in [t_0, t_1], N_t(f) = 1$; *elastic persistence*, where $Min_{t \in [t_0, t_1]} N_t(f) \in (0, 1)$; and *partially elastic persistence*, where $Min_{t \in [t_0, t_1]} N_t(f) = 0$. Elastic persistence should occur whenever the interval $[t_0, t_1]$ is short enough for the fluent to always remain somewhat certain; if the interval is too long, then we only have partially elastic persistence, and there are some time points within the interval when it cannot be guaranteed that the fluent is still somewhat cer-tain. Consider for example the fluent $free$ (again the parking place); if it is known that $free$ holds at 10.00 and at 10.10, nothing being about its truth value inbetween, it is reasonable to consider the case of elastic persistence (for it is almost certain that the place has remained free for the whole interval); now, if it is known that $free$ holds January 1st at 10.00 and at May 1st at 10.00, nothing being known about its truth value inbetween, then it is of course not reasonable to assume the same, since for time points far from both January 1st 10.00 and May 1st 10.00 it should be absolutely not certain that $free$ still holds.

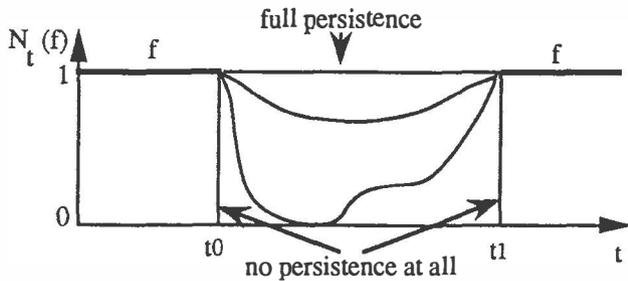

Figure 5: some functions for bounded extrapolation without change

*Bounded extrapolation with change*
Let $(H, f, (t_0, t_1))$ be a bounded extrapolation problem with change (without loss of generality, $f$ being True at $t_0$ and False at $t_1$). If we assume we do not want to generate partially inconsistent time-points (which is very reasonable), it must be always the case that $min(N_t(f), N_t(\neg f)) = 0$, thus the following axiom:
**D4.** $\exists t', t''$, with $t_0 \leq t' \leq t'' \leq t_1$ such that $N_t(f)$ is non-increasing in $[t_0, t']$, $N_t(f) = 0$ in $[t', t_1]$, $N_t(\neg f) = 0$ in $[t_0, t'']$ and $N_t(\neg f)$ is non-decreasing in $[t'', t_1]$.

### 4.2 Semi-quantitative axioms for decreasing persistence

The axioms we have given so far are very weak; in this subsection we give stronger axioms which do not use the metric properties of the certainty scale $[0, 1]$ but which use the metric properties of the temporal scale.

#### 4.2.1 Homogeneity

The main condition for a fluent being *homogeneous* is that the way it behaves with respect to decreasing persistence depends only on the *class* of the extrapolation problem and the time length of the interval, but not on *when* the interval starts. For instance, while the fluent "raining" may well be considered homogeneous on a time scale of 24 hours, it cannot be the case for the free parking place which will more certainly remain free after some period of time, say, at 10 pm than at 10 am. So, homogeneity should not always be required. However, in many cases, even if a fluent is definitely not homogeneous on the whole time scale, it can often be considered homogeneous on some shorter subintervals. The exact formulation of homogeneity is however more complex and expresses monotonicity conditions with respect to interval lengths. Let us now write formally some of the numerous homogeneity conditions. From now on, $f$ is a homogeneous fluent over the whole time scale.

*Case 1: monotonicity for two bounded extrapolation problems without change*
Let $H$ be a partial history; let $(H, f, (t_0, t_1))$ and $(H, f, (t_2, t_3))$ be two bounded extrapolation problems without change, the truth value of $f$ at the bounds of both intervals being identical (say, $True$). Homogeneity tells us that the shorter the interval, the more certain of the persistence of $f$ in the interval. For instance, if $free$ is homogeneous over $[8.00, 12.00]$, and is known to be true at 9.00, 9.10, 11.00 and 11.20, $free$ holding at 9.01 should be at least as certain than $free$ holding at 11.01, and similarly, $free$ holding at 9.09 should be at least as certain than $free$ holding at 11.19, for rather obvious reasons. Assume without loss of generality that $t_1 - t_0 \leq t_3 - t_2$, and let $\delta = t_1 - t_0$; then
**H1.** $\forall x \in [0, \delta], N_{t_0+x}(f) \leq N_{t_2+x}(f)$ and $\forall x \in [0, \delta], N_{t_1-x}(f) \leq N_{t_3-x}(f)$.
As an immediate consequence, if $t_1 - t_0 = t_3 - t_2$, then $\forall x \in [0, \delta], N_{t_0+x}(f) = N_{t_2+x}(f)$, i.e. the persistence function is exactly the same within two intervals of the same length.

*Case 2: monotonicity between forward extrapolation and bounded extrapolation without change*
Let $(H, f, (t_0, t_1))$ be a bounded extrapolation problem without change and $(H, f, (t_2, +\infty))$ be a forward extrapolation problem ($f$ being True at $t_0, t_1$ and $t_2$). Let $\delta = t_1 - t_0$. Then homogeneity tells that persistence should decrease at least as fast within $[t_3, +\infty)$ as in $[t_0, t_1]$, which writes
**H2.** $\forall x \in [0, \delta], N_{t_0+x}(f) \geq N_{t_2+x}(f)$.

*Case 3: bounded with change/ bounded with change*
Suppose we have two bounded extrapolation problems with change concerning the same fluent $f$, within the two intervals $(t_0, t_1)$ and $(t_2, t_3)$, the truth value of $f$ at $t_0$ and $t_2$ being the same (say, $True$). Then, homogeneity tells us that the shorter the interval, the faster persistence decreases $f$ in the interval (contrarily to what happens in the case of bounded persistence without change where the shorter the interval, the slower persistence decreases). Let us assume without loss of generality that $t_1 - t_0 \leq t_3 - t_2$ and let $\delta = t_1 - t_0$; then we get
**H3.** $\forall x \in [0, \delta], N_{t_0+x}(f) \geq N_{t_2+x}(f)$
and $\forall x \in [0, \delta], N_{t_1-x}(\neg f) \leq N_{t_3-x}(\neg f)$.

For the sake of brevity, we omit writing monotonicity conditions for the other cases (bounded without change/ backward, bounded with change/ bounded without change, bounded with change/ forward).

### 4.2.2  Other metric axioms

Among the other axioms we may require for some fluents, we can consider for instance *forward/backward symmetry*, which means that the fluent behaves symmetrically with respect to forward and backward extrapolation. Note that a lot of fluents don't (for instance, consider the well-known fluent *alive* of the Yale Shoooting Problem). Assuming both forward/backward symmetry and homogeneity for $f$ implies that backwards and forwards extrapolation functions are symmetric of each other, that functions for bounded persistence without change are symmetric relatively to the middle of the considered interval, and a symmetry property concerning bounded persistence. A stronger possible requirement (often too strong) is *symmetry with respect to negation*: the truth value "true" of the fluent tends to persist exactly the same way as the truth value "false". Among other things, it implies that, for a bounded persistence with change problem, the increasing functions for $f$ (resp. $\neg f$) and the increasing function for $\neg g$ (resp. $f$) are symmetric of each other.

### 4.3  Quantitative persistence functions

All the previous requirements do not enforce precise persistence functions. This last step (necessary for practical application) has to be done by the user. For instance, a reasonable choice for a family of persistence schemata consists in piecewise linear functions.

## 5  Inferring nonmonotonic conclusions from decreasing persistence

In Section 2, we have seen how, from a possibilistic knowledge base, it is possible to define a nonmonotonic inference relation. So, since the application of decreasing persistence principles to a partial history gives us a possibilistic knowledge base, it is then possible to draw some non monotonic inferences. More formally, let $H$ be a partial history on a time scale $T$, and let $Pers$ be a set of persistence schemata for a subset of the fluents involved in $H$. Let $Apply(Pers, H)$ be the application of $Pers$ to $H$ as defined in Section 3. Now, for any t, let $N_t^*$ be the necessity measure obtained by the application of the principle of minimum specificity (as in Section 2) to $Apply(Pers, H)_t$. Then, for any $t \in T$, we can define the nonmonotonic inference relation $\mid\sim_t$ as in Section 2. Let us now give a detailed example.

**Example**
Let us consider two machines A and B which may be either working or in failure at any time point. Let $A$ and $B$ be propositional fluents, $A$ (resp. $B$) being true iff A (resp. B) is working. Both machines are considered equivalent with respect to persistence; furthermore we assume that $A$ and $B$ are homogeneous on T, and the persistence functions of $A$ and $\neg A$ are represented on figure 6 (the time unit being the day).

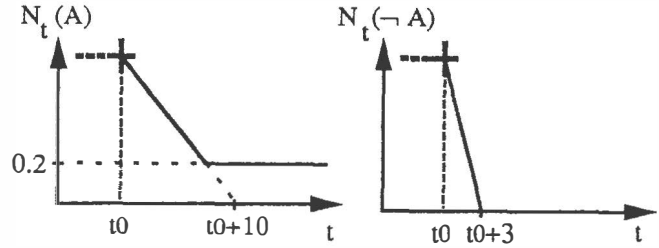

Figure 6:

Let us briefly comment these two persistence functions. The asymptotic value of 0.2 in the forward persistence function of *works* means that the certainty degree by default of *works* is 0.2, i.e. it is somewhat certain that machine work, independently from persistence considerations. The fastly decreasing persistence of $\neg works$ is due to the existence of repairmen (failing machines tend to be repaired in short delays). Let $K$ be the following timed knowledge base: machine A is known to be working from 0 to 10, machine B is known to be working from 17 to 30 and we know that at least one of the two machines is not in a failure state at time 15; formally:
$K = \{[0,10]: A; [15]: \neg A \vee \neg B; [17,30]: B\}$.
Now, consider the fluent $A$. We have successively a backward extrapolation problem on $(-\infty, 0)$, and then a forward extrapolation problem on $(10, +\infty)$. Applying the decreasing persistence schemata, we get the following certainty degrees at time 15: $N_{15}^*(A) = 0.5$; $N_{15}^*(B) = 0.8$ and (without needing persistence schemata) $N_{15}^*(\neg A \vee \neg B) = 1$. We have also $N_{15}^*(\neg A) = min(N_{15}^*(\neg A \vee \neg B), N_{15}^*(B)) = 0.8$. Moreover we get $N_{15}^*(\bot) = min(N_{15}^*(A), N_{15}^*(B), N_{15}^*(\neg A \vee \neg B)) = 0.5$; hence, the knowledge has an inconsistency degree at time 15.
Since $N_{15}^*(\neg A) = 0.8 > N_{15}^*(\bot)$, we have $\mid\sim_{15} \neg A$; similarly we have $\mid\sim_{15} B$, but we do not have $\mid\sim_{15} A$ or $\mid\sim_{15} \neg B$. This is due to the fact that the closest time point when $B$ is true is closer to 15 than the closest time point where $A$ is true.
Note also that at $t = 35$, we have $N_{35}^*(B) = 0.5$ and $N_{35}^*(\bot) = 0$, so we have $\mid\sim_{35} B$.

## 6  Concluding remarks

In this paper we have shown that possibility theory is well-suited for modelling gradually decreasing persistence, mainly because it is adequate to representing states of partial or complete ignorance. Moreover, since necessity orderings and similar constructions have been proved to be well-suited for performing nonmonotonic deductions, this framework provides us with a general methodology for inferring uncertain, defeasible conclusions from a "hard facts" knowledge base and some persistence schemata describing, for each fluent, how ignorance increases with respect to its persistence.

We think of pursuing our work in many directions. First of all, in this paper we considered decreasing persistence schemata only for *atomic* fluents; this leads to some problems when only disjunctions of fluents are known (see (Schrag 92) for a study of problems created by disjunction in reasoning about persistence). For instance, consider the partial history where $f \vee g$ is True at $t_0$, nothing else being known. Since both fluents $f$ and $g$ have the Unknown status at $t_0$, we can apply persistence schemata to none of them; and since there is no persistence schema for $f \vee g$, we will get $N_{t_0+\epsilon}(f \vee g) = 0 \ \forall \epsilon > 0$, i.e. no persistence at all for $f \vee g$. This could be avoided by applying persistence to non-atomic formulas as well; however, this leads to many technical problems, because persistence schemata of different formulas sharing fluents obviously interact. This is a topic left for further research.

We could also generalize our study to non-propositional fluents (i.e. whose domain is not True, False), which should not cause any trouble; we also think to incorporate decreasing persistence principles with non-gradual approaches dealing not only with persistence but more generally with time and action, such as in (Sandewall 92). Another easy generalisation of our work would consist in starting from a timed knowledge base already pervaded with uncertainty (i.e. from a possibilistic knowledge base) and to extrapolate necessity measures in a similar way. Moreover, work of Section 4 can be extended; in particular, it would be interesting to make a classification of fluents with respect to how they behave w.r.t decreasing persistence (adding other properties such as periodicity, ...).

Then, it would be interesting to generalize the principle of decreasing persistence to spatial reasoning (extrapolating the truth value of a fluent at a point $(x, y, z)$ by considering some close points where its truth value is known). Integrating both temporal and spatial "persistence" could enable us to infer defeasible conclusions from knowledge about time, space and motion. Next step would be a formal logical study of such a methodology, which could use notions of distances or similarity measures between worlds as in (Ruspini 91).

### Acknowledgements

We would like to thank Patrick Doherty, Didier Dubois and Henri Prade for helpful discussions. This work has been supported by the European ESPRIT Basic Research Action # 6156 entitled "Defeasible Reasoning and Uncertainty Management Systems (DRUMS-2)".